\documentclass{l4dc2027}

\title[BoBDeLaN]{BoBDeLaN: Bounded-Bandwidth Context Adaptation for Deep Lagrangian Robot Control}
\usepackage{times}
\usepackage{graphicx}
\usepackage{booktabs}

\newtheorem{assumption}{Assumption}

\newcommand{\equal}[1]{{\hypersetup{linkcolor=black}\thanks{#1}}}

\author{%
 \Name{Lucas Schulze}\equal{Equal Contribution.} \Email{lucas.schulze@robot-learning.de}\\
 \Name{Sucheth Shenoy}\footnotemark[1] \Email{sucheth@robot-learning.de}\\
 \Name{Jan Peters} \Email{jan@robot-learning.de}\\
 \Name{Oleg Arenz} \Email{oleg@robot-learning.de}\\
 \addr Department of Computer Science, Technical University of Darmstadt, Germany%
}

\newcommand{\Real}{\mathbb{R}}
\newcommand{\Transp}{^\mathrm{\scriptscriptstyle T}}

\newcommand{\jointTau}{\boldsymbol{\tau}}
\newcommand{\jointPos}{\mathbf{q}}
\newcommand{\jointVel}{\dot{\mathbf{q}}}
\newcommand{\jointAcc}{\ddot{\mathbf{q}}}

\newcommand{\PDKp}{\mathbf{K}_\mathrm{p}}
\newcommand{\PDKd}{\mathbf{K}_\mathrm{d}}

\newcommand{\Qmpc}{\mathbf{Q}}
\newcommand{\Rmpc}{\mathbf{R}}

\newcommand{\Ny}{\mathrm{n_y}}

\newcommand{\Lag}{\mathcal{L}}

\newcommand{\EPot}{P}
\newcommand{\inertiaMat}{\mathbf{H}}
\newcommand{\inertiaTri}{\mathbf{L}}

\newcommand{\latentVariableZ}{\mathbf{z}}
\newcommand{\tauChannelVar}{s}

\newcommand{\EoMC}{\mathbf{c}}
\newcommand{\EoMG}{\mathbf{g}}

\newcommand{\histInput}{\mathbf{h}}

\newcommand{\histLength}{\mathrm{n}_\mathrm{h}}

\newcommand{\jointTauNom}{\hat{\boldsymbol{\tau}}}

\newcommand{\jointTauDiffNom}{\tilde{\boldsymbol{\tau}}}

\newcommand{\KFQ}{\mathbf{Q}_\mathrm{EKF}}
\newcommand{\KFR}{\mathbf{R}_\mathrm{EKF}}

\newcommand{\massLoad}{\mathrm{m}_\mathrm{L}}

\newcommand{\posLoad}{\mathbf{r}_\mathrm{L}}

\newcommand{\DotLatentVarNorm}{\dot{\mathbf{z}}_\mathrm{n}}

\newcommand{\Nxi}{n_\xi}
\newcommand{\FrequencyGainMat}{\mathbf{G}}
\newcommand{\SinusoidalAmp}{a_s}

\begin{document}

\maketitle
{\renewcommand{\thefootnote}{}\renewcommand{\footnoteseptext}{}%
    \footnotetext{Project Page: \url{https://schulze18.github.io/bobdelan_website/}}}
\sloppy
\begin{abstract}
Learned physics-based models provide structure for robot control, but uncertain payloads and unmodeled dynamics limit their predictive accuracy. 
Context-aware Deep Lagrangian Networks enable adaptation through a latent context inferred from measurement histories while retaining Lagrangian structure for each fixed context. However, this structure alone does not constrain sensitivity to measurement disturbances. We introduce BoBDeLaN, which combines a contracting, gain-bounded Recurrent Equilibrium Network context encoder with stable linear filtering to shape disturbance sensitivity across frequencies. This architecture guarantees exponential forgetting of the recurrent initialization and admits a finite-horizon, frequency-weighted bound on context perturbations, which extends to residual torque prediction under a Lipschitz continuity assumption on the context-conditioned dynamics model. We evaluate a low-pass realization within model predictive control on simulated and real hardware manipulation tasks, together with frequency-sensitivity analysis on real-robot data. Compared with an LSTM encoder using the same input filter, BoBDeLaN exhibits reduced high-frequency context sensitivity, lower residual torque prediction error, and smoother motion, with less degradation under injected disturbances.
\end{abstract}

\begin{keywords}%
Deep Lagrangian Networks, Recurrent Equilibrium Networks, Adaptive Control, Frequency-Shaped Sensitivity, Robotic Manipulation
\end{keywords}



\subsection*{Generative AI Use}
Generative AI tools were used for coding assistance, plot generation, and proofreading. The authors take full responsibility for the accuracy and integrity of the submitted work.

\section{Introduction}

Physics-encoded learning incorporates mechanical structure directly into data-driven robot dynamics models~\citep{piccinini2026embeddingphysicspriorsrobot}. Approaches such as Hamiltonian Neural Networks (HNNs)~\citep{HNN} and Deep Lagrangian Neural Networks (DeLaN)~\citep{delan_lutter2019,delan_2021} exploit this structure to obtain physically consistent models. The resulting inductive bias is useful when data are limited~\citep{watson2025machinelearningphysicsknowledge} and supports integration with model-based control methods, such as model predictive control (MPC). In practice, however, friction, external disturbances, and payload variations can alter the dynamics, requiring adaptation to current operating conditions while preserving the model's mechanical structure.

Several works have explored this direction by combining physics encoded models with adaptive control. For example, \citet{Duong2021AdaptiveCO} proposed an adaptive energy controller based on HNNs with learned disturbance features, while \citet{delan_excavator} augmented DeLaN with convolutional and recurrent networks to model residual hydraulic actuation effects and friction. \cite{HU2026103292} combined DeLaN with adaptive Kalman filtering and noise suppression for contact force estimation. Context-Aware DeLaN~\citep{cadelac_iros} conditions a Lagrangian dynamics model on a latent context inferred from measurement histories by an LSTM encoder, enabling adaptation to unknown payloads within MPC. Although the model retains its mechanical structure for each fixed context, this structure does not constrain the sensitivity of the context encoder.

The measurement history contains information about changes in the dynamics together with measurement noise and estimation errors. Perturbations in this history can alter the inferred context and, consequently, the dynamics predictions supplied to MPC. Prediction accuracy on training data alone provides no guarantee on sensitivity to these perturbations or to recurrent state initialization. We therefore require exponential forgetting of the initial state under a common measurement history and bounded sensitivity of the context output to perturbations of that history. These requirements correspond to contraction and finite incremental gain, respectively, and are not guaranteed by standard recurrent architectures such as LSTMs.

Recurrent Equilibrium Networks (RENs) admit parametrizations that guarantee contraction and prescribed incremental integral quadratic constraints (IQCs), including incremental $\ell_2$ gain bounds, by construction~\citep{revay2024recurrent}. These guarantees hold throughout unconstrained optimization of the model parameters. More broadly, stable operator parametrizations have enabled learning-based controller design with closed loop stability guarantees, as demonstrated by Neural System Level Synthesis and magnitude and direction (MAD) policies~\citep{furieri2022neuralsls,furieri2025mad}. A uniform incremental gain bound, however, does not distinguish perturbations by temporal frequency. Frequency weighting refines this description by specifying how sensitivity varies with the frequency content of the perturbations. For REN based feedback control, \citet{barbara2025react} combine stable input and output filters with the recurrent parametrization to establish robustness guarantees. This construction motivates the composition of a stable linear filter and a certified recurrent model for context inference. The filter specifies the frequency weighting, while the incremental gain bound on the recurrent model yields a corresponding bound on context deviations.


In this work, we introduce BoBDeLaN, a frequency-shaped extension of Context-Aware DeLaN for adaptive robot dynamics learning with Lagrangian structure. We replace the LSTM encoder in Context-Aware DeLaN with a contracting, gain-bounded REN preceded by a low-pass filter. We establish theoretical bounds on context and residual torque prediction sensitivity.

The main contributions of this work are as follows. (i) A novel extension of Context-Aware DeLaN using a REN as the context encoder, enabling frequency-dependent context adaptation with contraction and incremental gain guarantees. (ii) We derive a finite-horizon frequency-weighted bound on context perturbations and, under a regularity assumption on the learned dynamics model, a corresponding bound on residual torque prediction variations along a fixed trajectory. (iii) We evaluate BoBDeLaN on real-robot data, in closed-loop simulation, and in a real hardware pick-and-place task, demonstrating reduced high-frequency context sensitivity, improved residual torque prediction accuracy, and smoother closed-loop behavior under payload variations and disturbances.

\section{Background}

\paragraph{Notation.}
We write $\mathbb N_0=\{0,1,\ldots\}$. 
The symbol $\|\cdot\|_2$ denotes the Euclidean norm for vectors and the induced spectral norm for matrices. We write $P\succeq0$ for a symmetric positive semidefinite matrix and define $\|v\|_P^2=v^\top Pv$. This defines a seminorm, which is a norm when $P\succ0$.
For a sequence $s=\{s_k\}_{k\in\mathbb N_0}$, define the truncated $\ell_2$-norm,
\begin{equation}
\label{eq:truncated-l2-norm}
    \|s\|_T:=\left(\sum_{k=0}^{T}\|s_k\|_2^2\right)^{1/2},
    \qquad T\in\mathbb N_0.
\end{equation}
We write $s\in\ell_2$ when $\|s\|_{\ell_2}:=\lim_{T\to\infty}\|s\|_T<\infty$. Finite-horizon inequalities do not require the full sequences to belong to $\ell_2$.  
For two sequences $s,s'$, we write $\Delta s=s-s'$, with $\Delta s_k=s_k-s'_k$. The symbols $\mathcal E,\mathcal R,\mathcal W$ denote operators on sequences.
Identity matrices $I$ have dimensions determined by context.

\subsection{Context-Aware Lagrangian Dynamics and Control}
\label{subsec:context-delan}

DeLaN~\citep{delan_lutter2019,delan_2021} models mechanical systems through a learned Lagrangian $\Lag(\jointPos,\jointVel)=\frac{1}{2}\jointVel\Transp\inertiaMat(\jointPos)\jointVel-\EPot(\jointPos)$, yielding the equations of motion via Euler–Lagrange
\begin{equation}
    \label{eq:eom_def}
    \inertiaMat(\jointPos)\jointAcc+\EoMC(\jointPos,\jointVel)+\EoMG(\jointPos)=\jointTau,
\end{equation}
where $\jointPos$ denotes joint positions, $\inertiaMat$ the inertia matrix, and $\EoMC$ and $\EoMG$ the Coriolis/centrifugal and gravitational torques. Two neural networks represent the inertia and potential energy, with positive definiteness ensured by parameterizing $\inertiaMat=\inertiaTri\inertiaTri\Transp$, where $\inertiaTri$ is a lower triangular matrix with positive diagonal entries. Differentiating the learned energy functions provides both inverse and forward dynamics from the same model.

For manipulation with unknown payloads, an available nominal robot model allows learning to focus on the residual torque $\jointTauDiffNom=\jointTau-\jointTauNom$. Context-Aware DeLaN~\citep{cadelac_iros} conditions the residual inertia and potential energy networks on a latent context $\latentVariableZ$, allowing a single model to represent different robot-payload combinations. 
%
%
In the original formulation, an LSTM \citep{hochreiter1997long} processes a history of length $\histLength$ with entries
\begin{equation}
    \label{eq:hist-input}
    \histInput_i=\begin{bmatrix}
        \jointPos_i\Transp & \jointVel_i\Transp & \jointTauDiffNom_i\Transp
    \end{bmatrix}\Transp.
\end{equation}
The joint states describe the observed motion, while the residual torques provide information about the mismatch with the nominal model. The encoder and Contextual DeLaN are trained jointly by minimizing inverse dynamics error. During deployment, the trained parameters remain fixed and adaptation occurs through the inferred context.
%
For each fixed $\latentVariableZ$, the energy derivatives are evaluated with respect to the mechanical coordinates while holding $\latentVariableZ$ constant. The inferred context may nevertheless change between evaluations as new measurements become available. This interpretation is useful when physical changes are infrequent, such as picking up or releasing an object, without requiring the context estimate itself to remain constant during adaptation.

Context-Aware Deep Lagrangian MPC (CaDeLaC)~\citep{cadelac_iros} uses the resulting model for joint trajectory tracking. Before each control update, the encoder estimates $\latentVariableZ_k$, which is then held fixed over the prediction horizon. The MPC combines nominal and learned dynamics to optimize tracking error and control effort subject to torque constraints, and applies the first control input. This allows the learned compensation to depend on the predicted motion throughout the horizon while updating the context at each control step.


\subsection{Contracting Recurrent Models}
\label{subsec:ren}

For recurrent context inference, we distinguish forgetting of the initial state from sensitivity to measurement perturbations. We consider causal recurrent models with state $\xi_k\in\Real^{\Nxi}$, input $\histInput_k^{\mathrm R}$, and context output $\latentVariableZ_k$. For a fixed parameter vector $\theta$ and initial state $\xi_0=a$, let $\mathcal R_\theta^a$ denote the induced input-output operator.

\begin{definition}[Contraction]
\label{def:contraction}
A recurrent model is contracting if there exist constants $K\geq1$ and $\rho\in(0,1)$ such that, for every common input sequence $\histInput^{\mathrm R}$ and every pair of initial states $a,b\in\Real^{\Nxi}$, the corresponding state trajectories satisfy
\begin{equation}
\label{eq:contraction}
    \|\xi_k^a-\xi_k^b\|_2
    \leq K\rho^k\|a-b\|_2,
    \qquad \forall k\in\mathbb N_0.
\end{equation}
\end{definition}

The constants are uniform over input sequences and initial states, but may depend on the model parameters. Contraction ensures exponential forgetting of the recurrent initialization under a common input, but does not itself quantify sensitivity to different inputs.

\begin{definition}[Incremental $\ell_2$ gain]
\label{def:incremental-gain}
A recurrent model has incremental $\ell_2$ gain at most $\gamma>0$ if, for every initial state $a$, every pair of input sequences $\histInput^{\mathrm R},{\histInput^{\mathrm R}}'$, and every $T\in\mathbb N_0$, the outputs $\latentVariableZ=\mathcal R_\theta^a(\histInput^{\mathrm R})$ and $\latentVariableZ'=\mathcal R_\theta^a({\histInput^{\mathrm R}}')$ satisfy
\begin{equation}
\label{eq:ren-l2-gain}
    \|\Delta\latentVariableZ\|_T
    \leq\gamma\|\Delta\histInput^{\mathrm R}\|_T.
\end{equation}
\end{definition}

This property bounds context perturbations induced by different input histories when the two trajectories have identical recurrent initial states.

\paragraph{Realization with RENs.}
Recurrent Equilibrium Networks (RENs) admit a direct parametrization of well-posed, contracting recurrent models with prescribed incremental input-output properties~\citep{revay2024recurrent}. For the gain-bounded RENs used here, the certificate also accounts for unequal initial states,
\begin{equation}
\label{eq:ren-iqc}
    \frac{1}{\gamma}\|\Delta\latentVariableZ\|_T^2
    \leq \gamma\|\Delta\histInput^{\mathrm R}\|_T^2+d(a,b),
    \qquad \forall T\in\mathbb N_0,
\end{equation}
where the two outputs are generated from recurrent initial states $a,b$. The initialization contribution $d(a,b)\geq0$ is finite, independent of the inputs and horizon, and satisfies $d(a,a)=0$. For identical recurrent initial states, $a=b$, the certificate recovers Definition~\ref{def:incremental-gain}. Inequality~\eqref{eq:ren-iqc} is an incremental integral quadratic constraint (IQC).

We use the acyclic REN realization, which permits sequential evaluation of its nonlinear variables, with element-wise activations whose slopes lie in $[0,1]$. Its direct parametrization enforces contraction and the incremental gain certificate for a prescribed $\gamma$ throughout unconstrained training, independently of optimizer convergence. We refer to~\citet{revay2024recurrent} for the construction.

\section{Frequency-Shaped Context Adaptation}
\label{sec:freq-shaped-context-adaptation}


In this section, we establish the context sensitivity bounds for the REN in BoBDeLaN and examine its implications for residual torque predictions. The low-pass filter determines the frequency weighting, while the REN certificate bounds the resulting sensitivity.

\subsection{Problem Formulation and Encoder Architecture}
\label{subsec:problem-formulation}

Our objective is a causal context encoder that forgets its recurrent initialization under a common measurement history and has controlled sensitivity to history perturbations. For payload adaptation, we favor slowly varying history features while attenuating fast measurement disturbances. This is a design preference: slowly changing payloads do not imply that all informative measurements are low-frequency. We use two cascaded first-order low-pass filters to strengthen attenuation of fast measurement disturbances while preserving unity steady-state gain. 

\begin{figure}[t]
    \centering
    \includegraphics{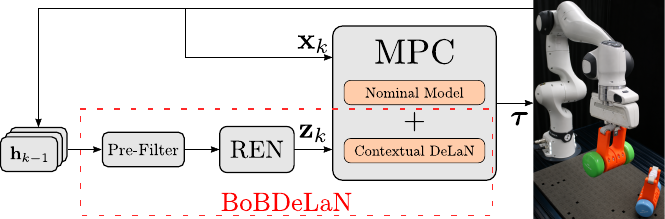}
    \caption{Context-Aware Deep Lagrangian Model Predictive Control~(CaDeLaC) using the BoBDeLaN (dashed red box). The Pre-Filter first normalizes and the filters the history $\histInput$. The MPC computes the torques $\jointTau$ from the state $\mathbf{x}_k$ using the nominal model and the contextual DeLaN. The REN can be replaced by any contracting encoder with bounded incremental gain.}
    \label{fig:bodelan_architecture}
\end{figure}

Let $\histInput$ denote the history sequence in~\eqref{eq:hist-input}, and let $\mathbf{D}_h$ be a fixed diagonal matrix of strictly positive channel-wise training standard deviations. We scale each history channel as
\begin{equation}
\label{eq:normalized-history}
    \bar{\histInput}_k=\mathbf{D}_h^{-1}\histInput_k.
\end{equation}
This scaling expresses the prescribed gain in normalized coordinates despite the different physical units of the history channels. The same filter is applied to every channel,
\begin{equation}
\label{eq:filtered-history-recursion}
    \begin{aligned}
    \mathbf{u}_k&=\alpha \mathbf{u}_{k-1}+(1-\alpha)\bar{\histInput}_k,&
    \mathbf{v}_k&=\alpha \mathbf{v}_{k-1}+(1-\alpha)\mathbf{u}_k,
    \qquad 0<\alpha<1,
    \end{aligned}
\end{equation}
where $\mathbf{u}_k$ is the first-stage output and $\mathbf{v}_k$ is supplied to the REN $\latentVariableZ=\mathcal R_\theta(\mathbf{v})$ satisfying~\eqref{eq:contraction} and~\eqref{eq:ren-iqc}. 
Writing $\mathcal W$ for the zero-state filter operator, the encoder with zero filter initialization is
\begin{equation}
\label{eq:frequency-weighted-observer-operator}
    \mathcal E_\theta
    =\mathcal R_\theta\circ\mathcal W\circ \mathbf{D}_h^{-1}.
\end{equation}
We suppress the dependence of $\mathcal E_\theta$ and $\mathcal R_\theta$ on their initial states in the notation. For nonzero filter initialization, we use~\eqref{eq:filtered-history-recursion} explicitly. Figure~\ref{fig:bodelan_architecture} depicts the proposed BoBDeLaN architecture with REN and the pre-filter integrated into CaDeLaC.
%
The scalar transfer function of the filter is
\begin{equation}
\label{eq:low-pass-filter}
    W(\zeta)=\left(\frac{1-\alpha}{1-\alpha\zeta^{-1}}\right)^2,
    \qquad
    \alpha=\exp\!\left(-\frac{2\pi f_b}{f_s}\right),
\end{equation}
where $f_s$ is the sampling frequency and $f_b$ is the per-stage pole-based bandwidth parameter. 
To characterize sensitivity, we compare paired encoder evaluations on a reference history $\histInput^0$ and a perturbed history
\begin{equation}
\label{eq:perturbed-history}
    \histInput=\histInput^0+\mathbf n,
    \qquad \bar{\mathbf n}_k=\mathbf{D}_h^{-1}\mathbf n_k,
\end{equation}
where $\mathbf n$ represents perturbations at the encoder input. With identical filter and recurrent initial states, the context difference is $\Delta\latentVariableZ=\mathcal E_\theta(\histInput^0+\mathbf n)-\mathcal E_\theta(\histInput^0)$. 
The construction also permits other stable causal linear prefilters, including band-pass, notch, and channel-dependent filters. Further extensions and the distinction between prefiltering and postfiltering are discussed in Appendix~\ref{app:input-output-shaping}.

\subsection{Frequency-Shaped Context Sensitivity}
\label{subsec:frequency-weighted-iqc}

The REN certificate yields a finite-horizon bound on context sensitivity after filtering and normalization.

\begin{proposition}[Context sensitivity under prefiltering]
\label{lem:frequency-shaped-iqc}
Suppose the recurrent model satisfies~\eqref{eq:ren-iqc}. For any pair of histories and identical filter and recurrent initial states, the context outputs satisfy
\begin{equation}
\label{eq:frequency-weighted-gain-result}
    \|\Delta\latentVariableZ\|_T
    \leq\gamma\|\mathcal W \mathbf{D}_h^{-1}\Delta\histInput\|_T,
    \qquad \forall T\in\mathbb N_0,
\end{equation}
where $\mathcal W$ acts on the input difference with zero initial states in both stages. More generally, let $a,b$ be the recurrent initial states, and let $\Delta \mathbf{u}_{-1}$ and $\Delta \mathbf{v}_{-1}$ be the differences in the two filter-stage initial states. The resulting zero-input output difference is
\(
    \mathbf{r}_k=\alpha^{k+1}\bigl[\Delta \mathbf{v}_{-1}+(k+1)(1-\alpha)\Delta \mathbf{u}_{-1}\bigr].
\) 
Then
\begin{equation}
\label{eq:frequency-weighted-iqc}
    \frac{1}{\gamma}\|\Delta\latentVariableZ\|_T^2
    \leq\gamma\|\mathcal W \mathbf{D}_h^{-1}\Delta\histInput+\mathbf{r}\|_T^2+d(a,b),
    \qquad \forall T\in\mathbb N_0.
\end{equation}
\end{proposition}
\begin{proof}
By linearity of the filter cascade,
$\Delta \mathbf{v}=\mathcal W \mathbf{D}_h^{-1}\Delta\histInput+\mathbf{r}$.
Solving the first-stage homogeneous recursion gives $\alpha^{k+1}\Delta \mathbf{u}_{-1}$; substitution into the second stage yields the stated $\mathbf{r}_k$. Substitution into~\eqref{eq:ren-iqc} gives~\eqref{eq:frequency-weighted-iqc}. For identical initial states, $\mathbf{r}=0$ and $d(a,a)=0$, yielding~\eqref{eq:frequency-weighted-gain-result}.
\end{proof}

Inequality~\eqref{eq:frequency-weighted-gain-result} expresses a frequency-shaped incremental gain bound, inherited from the REN certificate as a frequency-weighted IQC. It limits context deviations by the energy of the filtered, normalized history difference: the filter determines the frequency weighting, while $\gamma$ bounds the recurrent model's gain. For a common history and identical filter initial states, the REN receives identical inputs, so its recurrent state retains the exponential forgetting property in Definition~\ref{def:contraction}.

\paragraph{Frequency interpretation.}
The low-pass filter~\eqref{eq:low-pass-filter}, has unit gain at zero frequency and attenuates higher frequencies, thereby reducing the contribution of high-frequency disturbances to the energy bound.
This does not require the nonlinear encoder's empirical response to decrease monotonically with frequency. For the paired histories in~\eqref{eq:perturbed-history}, the finite-horizon bound~\eqref{eq:frequency-weighted-gain-result} applies with $\mathbf{D}_h^{-1}\Delta\histInput=\bar{\mathbf n}$. A spectral-energy interpretation is provided in Appendix~\ref{app:spectral-bound}.

\begin{corollary}[Sinusoidal disturbance bound]
\label{cor:sinusoidal-rms}
Let $\bar{\mathbf n}_k=A e_j\sin(\omega k+\varphi)$ for $k\geq0$, where $e_j$ is a unit coordinate vector, $A\geq0$, and $0<\omega<\pi$. Under the assumptions of Proposition~\ref{lem:frequency-shaped-iqc}, with any fixed finite initial states,
\begin{equation}
\label{eq:sinusoidal-rms-bound}
    \limsup_{T\to\infty}
    \frac{\|\Delta\latentVariableZ\|_T}{\sqrt{T+1}}
    \leq\frac{\gamma A}{\sqrt{2}}|W(e^{\mathrm i\omega})|.
\end{equation}
\end{corollary}
\begin{proof}
The filter response is a sinusoid of amplitude $A|W(e^{\mathrm i\omega})|$ plus an exponentially decaying transient. Its mean-square energy converges to $A^2|W(e^{\mathrm i\omega})|^2/2$. Divide~\eqref{eq:frequency-weighted-iqc} by $T+1$ and use the finite energy of $r$ and the finiteness of $d(a,b)$.
\end{proof}

In the experiments, a perturbation of amplitude $A$ times a channel's training standard deviation has normalized amplitude $A$, with $\omega=2\pi f/f_s$. The bound concerns the long-horizon RMS Euclidean norm of the full context perturbation, including harmonics. Singular values assembled from separate sinusoidal fits are empirical sensitivity statistics and are not directly certified by this result. For finite recordings with identical initial states, use~\eqref{eq:frequency-weighted-gain-result} with the actual filtered perturbation, retaining its transient, rather than treating the asymptotic RMS bound as a finite-window certificate. The attenuation--response-time trade-off of the prefilter is discussed in Appendix~\ref{app:attenuation-vs-response}.

\subsection{Sensitivity of Residual Torque Predictions}
\label{subsec:context-dynamics-robustness}

We compare residual torque predictions along the same motion trajectory, perturbing only the history supplied to the encoder. For brevity, write $x_k=(\jointPos_k,\jointVel_k,\jointAcc_k)$ and consider the inverse dynamics model from Context-Aware DeLaN as
\begin{equation}
\label{eq:context-residual-prediction}
    \widehat{\jointTauDiffNom}_k=\tilde{f}^{-1}(x_k,\latentVariableZ_k).
\end{equation}
Let $\widehat{\jointTauDiffNom}^{\,0}_k=\tilde{f}^{-1}(x_k,\latentVariableZ_k^0)$ be the prediction from the reference history and define $\Delta\widehat{\jointTauDiffNom}=\widehat{\jointTauDiffNom}-\widehat{\jointTauDiffNom}^{\,0}$. The direct inputs $x_k$ are identical in both evaluations, as in the offline disturbance-injection experiments.

\begin{assumption}[Lipschitz continuity in context]
\label{ass:context-regularity}
There exist sets $\mathcal X$, $\mathcal Z$ and a constant $0\leq L_z<\infty$ such that
\begin{equation}
\label{eq:context-lipschitz-bound}
    \|\tilde{f}^{-1}(x,z)-\tilde{f}^{-1}(x,z')\|_2
    \leq L_z\|z-z'\|_2
\end{equation}
for every $x\in\mathcal X$ and $z,z'\in\mathcal Z$. The compared direct inputs and both context trajectories remain in these sets over the evaluation horizon.
\end{assumption}
This condition concerns the Contextual DeLaN and is not enforced by the REN parametrization.

\begin{proposition}[Sensitivity of residual torque predictions]
\label{prop:residual-dynamics-robustness}
Suppose Assumption~\ref{ass:context-regularity} and Proposition~\ref{lem:frequency-shaped-iqc} apply. For the histories~\eqref{eq:perturbed-history}, identical filter and recurrent initial states, and identical direct inputs $x_k$,
\begin{equation}
\label{eq:residual-torque-robustness}
    \|\Delta\widehat{\jointTauDiffNom}\|_T
    \leq L_z\gamma\|\mathcal W\bar{\mathbf n}\|_T.
\end{equation}
\end{proposition}
\begin{proof}
Apply~\eqref{eq:context-lipschitz-bound} at each time, square and sum to obtain
$\|\Delta\widehat{\jointTauDiffNom}\|_T\leq L_z\|\Delta\latentVariableZ\|_T$.
Substitution of~\eqref{eq:frequency-weighted-gain-result} gives the result.
\end{proof}

The result bounds the additional variation in residual torque predictions caused by encoder-history disturbances along a fixed motion trajectory. By the triangle inequality, the disturbed prediction error is bounded by the reference prediction error plus this disturbance contribution. An explicit prediction-error decomposition and a sufficient Jacobian-based condition for context Lipschitz continuity are given in Appendix~\ref{app:context-regularity-details}. Numerical evaluation requires a valid $L_z$. The result does not establish closed-loop tracking or stability guarantees.

\section{Experiments}
We evaluate BoBDeLaN through frequency-sensitivity analysis on recorded Franka Panda data, simulated trajectory tracking, and hardware pick-and-place experiments. We compare against an LSTM encoder using the same input filter and, in closed-loop evaluations, nominal MPC. Dataset, training, and implementation details are provided in Appendix~\ref{sec:appendix_experimental_setup}.

\subsection{Frequency Sensitivity}
\label{subsec:exp-frequency-sensitivity}

We first validate the frequency-shaped context adaptation proposed in Sec.~\ref{sec:freq-shaped-context-adaptation} on real-robot data by injecting sinusoidal disturbances into individual joint torque inputs. We evaluate the resulting frequency sensitivity of the context and predicted torque for the LSTM and REN, and compare the REN response against its theoretical bounds (see Appendix~\ref{sec:appendix:exp-frequency-sensitivity} for details).

\begin{figure*}[h]
    \centering

    \subfigure[Largest singular value of the $\latentVariableZ$ response]{\includegraphics[width=208.14pt]{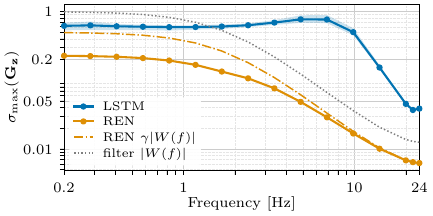}}
    \hfill
    \subfigure[RMS of the $\latentVariableZ$ response]{\includegraphics[width=208.14pt]{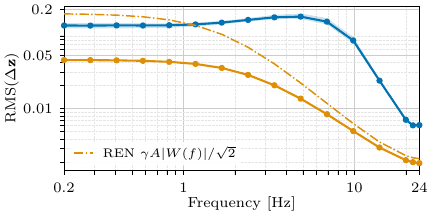}}\\
    \vspace{5pt}
    \subfigure[Largest singular value of the $\jointTau$ response]{\includegraphics[width=208.14pt]{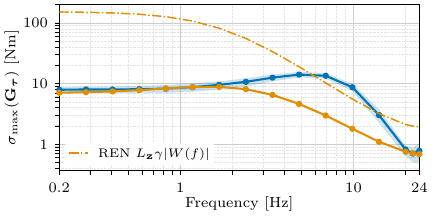}}
    \hfill
    \subfigure[RMS of the $\jointTau$ response]{\includegraphics[width=208.14pt]{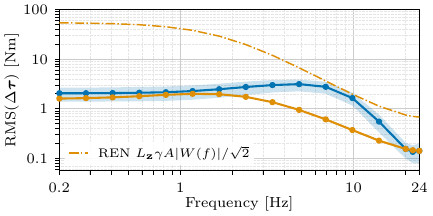}}
    \vspace{-5pt}
    \caption{Frequency sensitivity of the context $\latentVariableZ$ and predicted torque $\jointTau$ for the LSTM and REN, estimated over 12 real-robot runs. A sinusoidal disturbance with amplitude $A s_j$ and frequencies ranging from 0.2 to 24 Hz is applied to one joint torque input $j$ at a time, where $A = 0.5$ and $s_j$ denotes the standard deviation of the input. Dashed curves indicate REN reference curves ($\gamma = 0.5$), computed using the gain $|W(f)|$ of the second-order low-pass input filter at $f_b = 2~Hz$. The references on $\jointTau$ additionally account for $L_\mathbf{z}$, the maximum norm of $\partial\jointTau/\partial\latentVariableZ$ over the real-robot data. 
    }
    \label{fig:bode_all}
\end{figure*}
Figure~\ref{fig:bode_all}(a--b) reports the mean and interquartile range across runs of two context-sensitivity metrics: the largest singular value of the fitted fundamental-harmonic gain matrix and the full-response RMS, including harmonics. Both metrics exhibit high-frequency attenuation for the REN and a peak near 5~Hz for the LSTM under identical input filtering. The reference curve $\gamma|W(f)|$ follows from Proposition~\ref{lem:frequency-shaped-iqc} and bounds the induced frequency-response gain of the encoder linearized about a constant equilibrium, provided the linearization exists and is stable. This interpretation does not certify the fitted nonlinear gain matrices along the trajectories. Corollary~\ref{cor:sinusoidal-rms} provides the corresponding asymptotic RMS bound, rather than a finite-horizon guarantee.


Applying the same metrics to the predicted torque differences yields Fig.~\ref{fig:bode_all}(c--d). Both responses exhibit high-frequency attenuation for the REN, following a small peak near 1.5~Hz. With identical nominal model inputs, these differences coincide with the residual torque prediction differences bounded in~\eqref{eq:residual-torque-robustness}. The torque reference curves scale the corresponding context curves by the maximum sampled Jacobian norm $\|\partial\jointTau/\partial\latentVariableZ\|_2$. This sampled maximum does not establish a Lipschitz constant over the evaluation domain and the torque curves therefore remain empirical references.
\FloatBarrier

\subsection{Trajectory Tracking in Simulation}
\label{subsec:trajectory_trackin_sim}

We compare CaDeLaC with the prefiltered REN and LSTM encoders against nominal MPC without learned compensation. We use 30 unseen payload configurations sampled from the training distribution, with mass $\massLoad\sim\mathcal{U}(0,4)$~kg and each Cartesian component of $\posLoad$ sampled uniformly from $[-0.3,0.3]$~m. Evaluating 20 trajectories per payload yields 600 runs per method.


\begin{figure*}[h]
    \centering
    \includegraphics[width=0.8\linewidth]{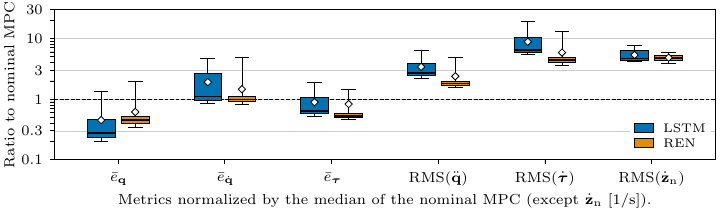}
    \caption{CaDeLaC results for REN and LSTM encoders in simulation for unseen payloads in the training distribution, 600 runs per model. Metrics are averaged over the 7 joints.}
    \label{fig:all_boxplot_id_results}
\end{figure*}
\begin{figure*}[h]
    \centering
    \includegraphics[width=0.8\linewidth]{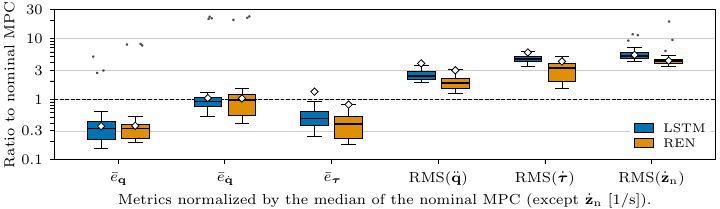}
    \caption{CaDeLaC results for REN and LSTM encoders in simulation for payloads out of training distribution ($\massLoad \sim \mathcal{U}(5,8)$~kg), 600 runs per model. Metrics are averaged over the 7 joints.} 
    \label{fig:all_boxplot_od_results}
\end{figure*}

Figure~\ref{fig:all_boxplot_id_results} reports tracking and smoothness metrics normalized by the corresponding nominal-MPC median, except the normalized context rate, which is reported directly. Position errors $\bar{e}_{\mathbf{q}}$ (joint-averaged RMS) are 45\% and 28\% of nominal, with the REN exhibiting half the LSTM's interquartile range.
Velocity errors $\bar{e}_{\dot{\mathbf{q}}}$ are comparable to nominal, with greater spread for the LSTM, while residual torque errors $\bar{e}_{\boldsymbol{\tau}}$ are reduced to 53\% and 64\%.
The REN yields smoother motion and control, with acceleration $\mathrm{RMS}$ of 1.8 versus 2.7 times nominal and torque rate $\mathrm{RMS}$ of 4.4 versus 6.4 times nominal. Normalized context rates $\mathrm{RMS}(\dot{\mathbf{z}}_\mathrm{n})$ are similar, with interquartile ranges of 0.9 and 2.0.
%
The REN generally exhibits narrower interquartile ranges.

To assess generalization beyond the training payload range, we evaluate 30 additional environments with out-of-distribution payload masses $\massLoad\sim\mathcal{U}(5,8)$~kg and torque limits increased fivefold. Figure~\ref{fig:all_boxplot_od_results} reports the same metrics. Median position errors are 33\% of nominal for both controllers, while velocity errors are similar, and residual torque errors are reduced to 39\% and 48\% for REN and LSTM, respectively. The REN remains smoother, with acceleration RMS of 184\% versus 238\% of nominal and torque-rate RMS of 328\% versus 455\%, accompanied by lower normalized context-rate RMS (4.23 versus 5.09). Each controller encounters QP failures in 3 of the 600 runs. The numerical values of all the metrics in both experiments are reported in Appendix~\ref{app:tracking_results}.

\subsection{Hardware Experiments}
We evaluate both controllers on a pick-and-place task on a real Franka Panda robot. The 26s trajectory involves four changes in system dynamics, as the robot successively picks up a 3~kg dumbbell and a 2~kg dumbbell. To evaluate the frequency sensitivity of both encoders, we apply sinusoidal disturbance signals with phase shifts to the first five joints at frequencies of 5~Hz and 10~Hz. Details and all numerical results are provided in Appendix~\ref{sec:appendix_pp_task_desc}. 
Videos are available on the project page.

\begin{figure*}[htbp]
    \centering
    \includegraphics[width=0.8\linewidth]{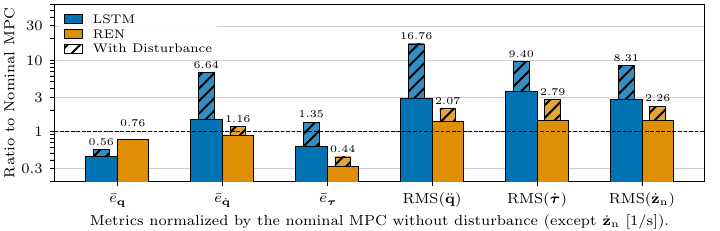}
    \caption{Pick-and-place task results for CaDeLaC with REN and LSTM encoders, evaluated with and without residual torque disturbance injection.}
    \label{fig:real_robot_dist_metrics}
\end{figure*}

Without injected disturbances, position errors are 78\% and 45\% of nominal MPC for REN and LSTM, respectively, while velocity errors are 88\% and 149\% and residual torque errors are reduced to 33\% and 62\% (Fig.~\ref{fig:real_robot_dist_metrics}). Although the LSTM achieves lower position error, the REN yields lower velocity and torque errors and smoother motion: acceleration and torque-rate RMS are both approximately 140\% of nominal for the REN, compared with 290\% and 360\% for the LSTM.

Under disturbance, the REN maintains its tracking and torque estimation performance, with only a moderate increase in velocity error from 88\% to 116\%. In contrast, the LSTM degrades across all metrics, with particularly pronounced increases in velocity and acceleration errors, reaching 665\% and 1680\%, respectively. This degradation is accompanied by an increase in normalized context-rate RMS $\DotLatentVarNorm$ from 2.83 to 8.31 for the LSTM, compared with 1.44 to 2.26 for the REN.

\section{Conclusion}

This work introduces BoBDeLaN, an extension of Context-Aware DeLaN combining a contracting, gain-bounded recurrent encoder with stable linear filtering for frequency-shaped context adaptation. We establish finite-horizon bounds on context perturbations and, under a context Lipschitz assumption, residual torque prediction variations along fixed trajectories. Experiments on real-robot data, closed-loop simulation, and hardware pick-and-place demonstrate reduced high-frequency context sensitivity, lower residual torque prediction error, and smoother motion than an LSTM encoder using the same input filter.


Future work will explore alternative stable linear filters to shape BoBDeLaN's sensitivity across different bandwidths, enabling frequency-selective adaptation and improved robustness to effects such as friction and actuator dynamics. Another direction is extending it to physics-encoded learned dynamics models for different robotic systems, addressing application-specific disturbances such as aerodynamic effects in quadrotors~\citep{port_hnn} and contact-related disturbances in legged robots~\citep{felan} and exoskeletons~\citep{pilan, exolan}.

\bibliography{l4dc2027-sample}

@INPROCEEDINGS{cadelac_iros,
  author={Schulze, Lucas and Peters, Jan and Arenz, Oleg},
  booktitle={2025 IEEE/RSJ International Conference on Intelligent Robots and Systems (IROS)}, 
  title={Context-Aware Deep Lagrangian Networks for Model Predictive Control}, 
  year={2025},
  volume={},
  number={},
  pages={6939-6946},
  doi={10.1109/IROS60139.2025.11246292}}

@article{delan_2021,
  author={Michael Lutter and Jan Peters},
  title={Combining Physics and Deep Learning to learn Continuous-Time Dynamics Models},
  journal={CoRR},
  volume={abs/2110.01894},
  year={2021},
  url={},
  eprinttype={arXiv},
  eprint={2110.01894},
  bibsource={dblp computer science bibliography, https://dblp.org}
}

@inproceedings{
    delan_lutter2019,
    title={Deep Lagrangian Networks: Using Physics as Model Prior for Deep Learning},
    author={Michael Lutter and others},
    booktitle={International Conference on Learning Representations},
    year={2019},
}

@article{hochreiter1997long,
  title={Long short-term memory},
  author={Hochreiter, Sepp and Schmidhuber, J{\"u}rgen},
  journal={Neural computation},
  volume={9},
  number={8},
  pages={1735--1780},
  year={1997},
  publisher={MIT press}
}

@article{revay2024recurrent,
  author  = {Revay, Max and Wang, Ruigang and Manchester, Ian R.},
  title   = {Recurrent Equilibrium Networks: Flexible Dynamic Models with Guaranteed Stability and Robustness},
  journal = {IEEE Transactions on Automatic Control},
  volume  = {69},
  number  = {5},
  pages   = {2855--2870},
  year    = {2024},
  doi     = {10.1109/TAC.2023.3294101}
}

@misc{piccinini2026embeddingphysicspriorsrobot,
      title={Embedding Physics Priors in Robot Learning: A Survey}, 
      author={Mattia Piccinini and Lucas Schulze and Alice Plebe and Matteo Saveriano and Thomas Beckers and Yuan Gao and Oleg Arenz and Baha Zarrouki and Dingrui Wang and Finn Rasmus Schäfer and Jan Peters and Johannes Betz and Gastone Pietro Rosati Papini},
      year={2026},
      eprint={2609.22319},
      archivePrefix={arXiv},
      primaryClass={cs.RO},
      url={https://arxiv.org/abs/2609.22319}, 
}

@article{
watson2025machinelearningphysicsknowledge,
title={Machine Learning with Physics Knowledge for Prediction: A Survey},
author={Joe Watson and Chen Song and Oliver Weeger and Theo Gruner and An Thai Le and Kay Hansel and Ahmed Hendawy and Oleg Arenz and Will Trojak and Miles Cranmer and Carlo D'Eramo and Fabian Buelow and Tanmay Goyal and Jan Peters and Martin W Hoffmann},
journal={Transactions on Machine Learning Research},
issn={2835-8856},
year={2025},
url={https://openreview.net/forum?id=ZiJYahyXLU},
note={Survey Certification}
}

@inproceedings{HNN,
 author = {Greydanus, Samuel and Dzamba, Misko and Yosinski, Jason},
 booktitle = {Advances in Neural Information Processing Systems},
 publisher = {Curran Associates, Inc.},
 title = {Hamiltonian Neural Networks},
 volume = {32},
 year = {2019}
}

@inproceedings{todorov2012mujoco,
  title={MuJoCo: A physics engine for model-based control},
  author={Todorov, Emanuel and Erez, Tom and Tassa, Yuval},
  booktitle={2012 IEEE/RSJ International Conference on Intelligent Robots and Systems},
  pages={5026--5033},
  year={2012},
  organization={IEEE},
  doi={10.1109/IROS.2012.6386109}
}

@Article{acados_2021,
  Title={acados -- a modular open-source framework for fast embedded optimal control},
  Author={Robin Verschueren and others},
  Journal={Mathematical Programming Computation},
  Year={2021},
}

@article{l4_casadi_taylor_salzmann2023neural,
  title={Real-time Neural-MPC: Deep Learning Model Predictive Control for Quadrotors and Agile Robotic Platforms},
  author={Salzmann, Tim and Kaufmann, Elia and Arrizabalaga, Jon and Pavone, Marco and Scaramuzza, Davide and Ryll, Markus},
  journal={IEEE Robotics and Automation Letters},
  doi={10.1109/LRA.2023.3246839},
  year={2023}
}

@inproceedings{carpentier2019pinocchio,
   title={The Pinocchio C++ library -- A fast and flexible implementation of rigid body dynamics algorithms and their analytical derivatives},
   author={Carpentier, Justin and others},
   booktitle={IEEE International Symposium on System Integrations (SII)},
   year={2019}
}

@article{HPIPM,
    title = {{HPIPM: a high-performance quadratic programming framework for model predictive control}},
    journal = {IFAC-PapersOnLine},
    volume = {53},
    number = {2},
    pages = {6563-6569},
    year = {2020},
    note = {21st IFAC World Congress},
    issn = {2405-8963},
    author = {Gianluca Frison and Moritz Diehl},
}

@misc{barbara2025react,
  title         = {React to Surprises: Stable-by-Design Neural
                   Feedback Control and the {Youla-REN}},
  author        = {Barbara, Nicholas H. and Wang, Ruigang
                   and Megretski, Alexandre and Manchester, Ian R.},
  year          = {2025},
  eprint        = {2506.01226},
  archivePrefix = {arXiv},
  primaryClass  = {eess.SY},
  url           = {https://arxiv.org/abs/2506.01226}
}

@article{Duong2021AdaptiveCO,
  title={Adaptive Control of SE(3) Hamiltonian Dynamics With Learned Disturbance Features},
  author={Thai Duong and Nikolay A. Atanasov},
  journal={IEEE Control Systems Letters},
  year={2021},
  volume={6},
  pages={2773-2778},
  url={https://api.semanticscholar.org/CorpusID:247597120}
}

@article{delan_excavator,
title = {Physics-informed neutral network with physically consistent and residual learning for excavator precision operation control},
journal = {Applied Soft Computing},
volume = {167},
pages = {112402},
year = {2024},
issn = {1568-4946},
doi = {https://doi.org/10.1016/j.asoc.2024.112402},
url = {https://www.sciencedirect.com/science/article/pii/S1568494624011761},
author = {Chenlong Feng and Jixin Wang and Yuying Shen and Qi Wang and Yi Xiong and Xudong Zhang and Jiuchen Fan},
}

@article{HU2026103292,
title = {Physics-informed adaptive Kalman filter for contact force estimation in industrial robots considering model uncertainty},
journal = {Robotics and Computer-Integrated Manufacturing},
volume = {101},
pages = {103292},
year = {2026},
issn = {0736-5845},
doi = {https://doi.org/10.1016/j.rcim.2026.103292},
url = {https://www.sciencedirect.com/science/article/pii/S0736584526000712},
author = {Hongbo Hu and Jia Ma and Yuhan Lu and Chungang Zhuang},
}

@INPROCEEDINGS{furieri2025mad,
  author={Furieri, Luca and Shenoy, Sucheth and Saccani, Danilo and Martin, Andrea and Ferrari-Trecate, Giancarlo},
  booktitle={2025 IEEE 64th Conference on Decision and Control (CDC)}, 
  title={MAD: A Magnitude And Direction Policy Parametrization for Stability Constrained Reinforcement Learning}, 
  year={2025},
  volume={},
  number={},
  pages={942-947},
  doi={10.1109/CDC57313.2025.11312015}}

@INPROCEEDINGS{furieri2022neuralsls,
  author={Furieri, Luca and Galimberti, Clara Lucía and Ferrari-Trecate, Giancarlo},
  booktitle={2022 IEEE 61st Conference on Decision and Control (CDC)}, 
  title={Neural System Level Synthesis: Learning over All Stabilizing Policies for Nonlinear Systems}, 
  year={2022},
  volume={},
  number={},
  pages={2765-2770},
  doi={10.1109/CDC51059.2022.9992492}}

@INPROCEEDINGS{felan,
  author={Schulze, Lucas and Negri, Juliano Decico and Barasuol, Victor and Medeiros, Vivian Suzano and Becker, Marcelo and Peters, Jan and Arenz, Oleg},
  booktitle={2026 IEEE International Conference on Robotics and Automation (ICRA)}, 
  title={Floating-Base Deep Lagrangian Networks}, 
  year={2026},
  volume={},
  number={},
  pages={18554-18561},
  doi={10.1109/ICRA57385.2026.11696543}
}

@misc{exolan,
      title={{ExoLaN}: Physics-Consistent Context-Aware Dynamics Learning for Exoskeletons}, 
      author={Lucas Schulze and Maximilian Schwarz and Jona Hoppe and Jan Peters and Oleg Arenz},
      year={2026},
      eprint={2609.31434},
      archivePrefix={arXiv},
      primaryClass={cs.RO},
      url={https://arxiv.org/abs/2609.31434}, 
}

@ARTICLE{pilan,
  author={Chen, Longwen and Cui, Fangge and Lu, Huimin},
  journal={IEEE Robotics and Automation Letters}, 
  title={PILaN: Generating Task-Individual Independent Customized Assistive Control on a Hip-Knee Powered Exoskeleton}, 
  year={2026},
  volume={11},
  number={5},
  pages={5382-5389},
  doi={10.1109/LRA.2026.3666397}
}

@ARTICLE{port_hnn,
  author={Duong, Thai and Altawaitan, Abdullah and Stanley, Jason and Atanasov, Nikolay},
  journal={IEEE Transactions on Robotics}, 
  title={Port-Hamiltonian Neural ODE Networks on Lie Groups for Robot Dynamics Learning and Control}, 
  year={2024},
  volume={40},
  number={},
  pages={3695-3715},
  doi={10.1109/TRO.2024.3428433}}

\newpage
\appendix
\section{Extensions to Input and Output Frequency Shaping}
\label{app:input-output-shaping}
\begin{remark}[Input and output frequency shaping]
\label{rem:general-frequency-shaping}
The construction extends to stable causal linear time-invariant prefilters and postfilters of compatible dimensions,
\begin{equation}
\label{eq:general-filtered-encoder}
    \mathcal E_\theta
    =\mathcal W_{\mathrm{out}}\circ\mathcal R_\theta
       \circ\mathcal W_{\mathrm{in}}\circ \mathbf{D}_h^{-1}.
\end{equation}
This is the same input/output weighting structure used for the learnable recurrent operator in~\citet[Sec.~V-E]{barbara2025react}. Taking either filter to be the identity gives input-only or output-only shaping. The prefilter shapes disturbances entering the recurrent model, the postfilter acts on the resulting context signal, including frequency components generated by the nonlinear recurrent dynamics. Stable band-pass filters can emphasize selected frequency ranges, notch filters can attenuate narrow frequency bands, and channel-dependent filters can encode different sensitivity requirements. Since the recurrent model is nonlinear, the two placements are generally not interchangeable, and their frequency responses cannot simply be multiplied to obtain a pointwise encoder gain bound. The experiments in this paper use $\mathcal W_{\mathrm{in}}=\mathcal W$ and $\mathcal W_{\mathrm{out}}=I$.
\end{remark}

\section{Spectral Interpretation of the Context Bound}
\label{app:spectral-bound}
For a history perturbation $\mathbf n$, let $\bar{\mathbf n}=\mathbf{D}_h^{-1}\mathbf n$ denote its channel-wise normalized value. 
The encoder consists of a stable linear filter $\mathcal W$ followed by a contracting recurrent model $\mathcal R_\theta$ with incremental gain at most $\gamma$. 
For paired encoder evaluations with identical filter and recurrent initial states, the main paper establishes
\begin{equation}
\label{eq:supp-finite-gain}
\|\Delta\latentVariableZ\|_T\leq\gamma\|\mathcal W\bar{\mathbf n}\|_T,
\qquad T\in\mathbb N_0.
\end{equation}
Here $\Delta\latentVariableZ$ is the difference between the disturbed and reference context sequences, and $\mathcal W$ acts on the perturbation with zero initial state.

\begin{corollary}[Spectral energy bound]
\label{cor:spectral-energy}
Suppose $\bar{\mathbf n}\in\ell_2$ and the filter and recurrent initial states are identical. Then $\Delta\latentVariableZ\in\ell_2$ and
\begin{equation}
\label{eq:frequency-domain-inequality}
    \|\Delta\latentVariableZ\|_{\ell_2}^2
    \leq\frac{\gamma^2}{2\pi}
       \int_{-\pi}^{\pi}|W(e^{\mathrm i\omega})|^2
       \|\widehat{\bar{\mathbf n}}(\omega)\|_2^2\,\mathrm d\omega,
\end{equation}
where the hat denotes the discrete-time Fourier transform of the sequence extended by zero to negative indices.
\end{corollary}
\begin{proof}
Stability of $\mathcal W$ implies $\mathcal W\bar{\mathbf n}\in\ell_2$. Taking $T\to\infty$ in~\eqref{eq:supp-finite-gain} bounds the output energy. Parseval's identity gives~\eqref{eq:frequency-domain-inequality}.
\end{proof}

\section{Attenuation and Response Time}
\label{app:attenuation-vs-response}
The prefilter consists of two identical first-order stages,
$\mathbf{u}_k=\alpha \mathbf{u}_{k-1}+(1-\alpha)\bar{\histInput}_k$ and
$\mathbf{v}_k=\alpha \mathbf{v}_{k-1}+(1-\alpha)\mathbf{u}_k$.
Consider a step in the normalized history from $\bar{\histInput}_-$ to $\bar{\histInput}_+$ at $k=0$, with both stages initially at steady state, $\mathbf{u}_{-1}=\mathbf{v}_{-1}=\bar{\histInput}_-$. Solving the first-stage recursion and substituting its response into the second stage gives
\begin{equation}
\label{app:eq:filter-step-response}
    \mathbf{v}_k-\bar{\histInput}_+=\alpha^{k+1}\bigl[1+(k+1)(1-\alpha)\bigr](\bar{\histInput}_--\bar{\histInput}_+),
    \qquad k\geq0.
\end{equation}
Compared with a single stage using the same $\alpha$, the cascade squares the frequency-response magnitude and therefore strengthens attenuation at nonzero frequencies. The additional factor $1+(k+1)(1-\alpha)$ in the step-response error makes the response slower. Increasing $\alpha$ strengthens attenuation at nonzero frequencies but slows the decay of the step-response error, delaying changes presented to the REN. With $\alpha=\exp(-2\pi f_b/f_s)$, where $f_s$ is the sampling frequency and $f_b$ the per-stage pole-based bandwidth parameter, a smaller $f_b$ produces stronger attenuation and a slower response. Equation~\eqref{app:eq:filter-step-response} characterizes the prefilter response, not the time required to identify a new payload. The latter also depends on the observed motion and the learned encoder and dynamics model.

\section{Context Regularity and Prediction Error}
\label{app:context-regularity-details}
\label{sec:context-regularity-details}
Write $x=(\jointPos,\jointVel,\jointAcc)$ and let $\tilde{f}^{-1}(x,z)$ denote the residual inverse dynamics model with its trained network parameters held fixed.

\paragraph{A sufficient condition for context Lipschitz continuity.}
Suppose $\mathcal X\times\mathcal Z$ is compact, $\mathcal Z$ is convex, and the context Jacobian of $\tilde{f}^{-1}$ exists and is continuous on a neighborhood of this product set. Then a finite Lipschitz constant is
\begin{equation}
\label{eq:context-jacobian-bound}
    L_z=\sup_{(x,z)\in\mathcal X\times\mathcal Z}
           \left\|\frac{\partial \tilde{f}^{-1}}{\partial z}(x,z)\right\|_2.
\end{equation}
Indeed, for fixed $x$, integrate the context Jacobian along the segment from $z'$ to $z$. Convexity keeps this segment in $\mathcal Z$, giving
$\|\tilde{f}^{-1}(x,z)-\tilde{f}^{-1}(x,z')\|_2\leq L_z\|z-z'\|_2$.
Continuity and compactness ensure that the supremum is finite. Both compared context trajectories and the direct inputs must remain in these sets. A maximum over sampled Jacobians is only an empirical sensitivity estimate and does not establish an upper bound over the entire domain.

\paragraph{Prediction-error decomposition.}
Consider histories $\histInput^0$ and $\histInput^0+\mathbf n$, with normalized difference $\bar{\mathbf n}=\mathbf{D}_h^{-1}\mathbf n$, identical filter and recurrent initial states, and identical direct inputs $x_k$. Let $\widehat{\jointTauDiffNom}^{\,0}$ and $\widehat{\jointTauDiffNom}$ denote the resulting reference and disturbed predictions. Under the context Lipschitz condition, the encoder bound~\eqref{eq:supp-finite-gain} gives
$\|\widehat{\jointTauDiffNom}-\widehat{\jointTauDiffNom}^{\,0}\|_T\leq L_z\gamma\|\mathcal W\bar{\mathbf n}\|_T$.
For a target residual torque sequence $\jointTauDiffNom^\star$, define the reference prediction error
$\varepsilon_T^0=\|\jointTauDiffNom^\star-\widehat{\jointTauDiffNom}^{\,0}\|_T$.
The triangle inequality then yields
\begin{equation}
\label{eq:residual-error-decomposition}
    \|\jointTauDiffNom^\star-\widehat{\jointTauDiffNom}\|_T
    \leq\varepsilon_T^0+L_z\gamma\|\mathcal W\bar{\mathbf n}\|_T.
\end{equation}
This separates the reference prediction error from the additional contribution due to perturbations of the encoder history; it does not bound the reference error itself.

\section{Experimental Setup}
\label{sec:appendix_experimental_setup}

\subsection{Dataset and Training Details}

We use the dataset provided by \citep{cadelac_iros}, collected in MuJoCo~\citep{todorov2012mujoco}. It comprises 2020 trajectories of 10~s each, sampled at 50~Hz, covering 100 randomly generated payload configurations. Payload masses $\massLoad$ are uniformly sampled from 0 to 4~kg, and payload positions $\posLoad$ from $[-0.3\text{m}, 0.3\text{m}]$ along each Cartesian axis in the end-effector frame. The dataset also includes trajectories without additional payload. Further details on the dataset collection can be found in~\citep{cadelac_iros}.

Each MLP in the Contextual DeLaN consists of two hidden layers with 30 and 20 neurons, respectively. The context encoder is an LSTM with five hidden layers of 10 units each, followed by a dense layer applied to the final temporal output to produce a 10-dimensional latent vector. The REN has 10-dimensional state and nonlinear input vectors and is constrained to an incremental $\ell_2$-gain of $\gamma = 0.5$. The historical input, defined in \eqref{eq:hist-input}, has 21 dimensions. All networks are trained for 1500 epochs using PyTorch with five random seeds. For the frequency sensitivity analysis and control experiments, we select the model achieving the best performance on the test set. Training hyperparameters are presented in Tab.~\ref{tab:train_parameters}.

\begin{table}[ht]
\centering
\caption{Training hyperparameters.}
\begin{tabular}{cc}
\toprule
\textbf{Parameter} & \textbf{Values} \\
\toprule
Activation & Tanh\\
Batch Size & 2048\\
Learning Rate & $5 \times 10^{-4}$\\
Epochs & 1500\\
$\histLength$ & 15 \\
$\epsilon$ & 0.1 \\
\bottomrule
\end{tabular}
\label{tab:train_parameters}
\end{table}

\subsection{Control Implementation}

We employ the nonlinear MPC implementation from \citep{cadelac_iros}, based on Acados~\citep{acados_2021} in SQP-RTI mode with the HPIPM solver~\citep{HPIPM}. The nominal model is implemented as a CasADi expression using Pinocchio~\citep{carpentier2019pinocchio}. To convert the PyTorch models into symbolic expressions, we use L4CasADi~\citep{l4_casadi_taylor_salzmann2023neural} to integrate the DeLaN networks into the MPC. Controllers parameters are presented in Tab.~\ref{tab:control_parameters}. 

\begin{table}[ht]
\centering
\caption{Controller parameters.}
\begin{tabular}{cc}
\toprule
\textbf{Parameter} & \textbf{Values} \\
\toprule
$\Ny$ & 12 \\
Control frequency & 50~Hz \\
Discretization time & 20~ms \\
$\Qmpc$ & $\mathrm{diag}(${\scriptsize$100 \times\mathbf{1}_{4}, 50 \times\mathbf{1}_{3}, 10^{3} \times \mathbf{1}_{4}, 500 \times\mathbf{1}_{3}$}$)$ \\
$\Rmpc$ & $\mathrm{diag}(0.2 \times\mathbf{1}_{4}, 1.0 \times \mathbf{1}_{3})$ \\
$\jointTau_\mathrm{max}$, $-\jointTau_\mathrm{min}$ & [87, 87, 87, 87, 12, 12, 12] \\
$\KFQ$ & $\mathrm{diag}(\mathbf{1}_{7}, 10^{-1} \times \mathbf{1}_{7}, 10^{2} \times\mathbf{1}_{3})$ \\
$\KFR$ & $\mathrm{diag}(10^{-2} \times \mathbf{1}_{14})$ \\
$\PDKp$ & [10, 10, 10, 10, 5, 5, 2] \\
$\PDKd$ & [2.5, 2.5, 2.5, 2.5, 1.5, 1.5, 1.5] \\
\bottomrule
\end{tabular}
\label{tab:control_parameters}
\end{table}

\section{Frequency Sensitivity Computation}
\label{sec:appendix:exp-frequency-sensitivity}

We evaluate the frequency-shaped context adaptation introduced in Sec.~\ref{sec:freq-shaped-context-adaptation} using real-robot data from a Franka Panda performing manipulation tasks. The dataset comprises an infinity-symbol trajectory with payloads of 1, 2, and 3~kg and a pick-and-place trajectory with payloads of 2 and 3~kg. Each configuration is executed using three controllers from \citep{cadelac_iros}: the nominal, Extended Kalman Filter (EKF), and CaDeLaC controllers, resulting in 12 runs in total. The runs last between 21.6~s and 27~s and are sampled at 50~Hz. Following training, the encoder input is low-pass filtered at 2~Hz and normalized using the training standard deviation $\tauChannelVar_j$ of each channel.

To measure how a disturbance of the residual torque propagates, we add a sinusoid $\SinusoidalAmp\tauChannelVar_j\sin(2\pi ft)$ to channel $j$ of $\jointTauDiffNom$ before filtering, one joint at a time with $\SinusoidalAmp = 0.5$, while the remaining encoder and DeLaN inputs are left unchanged. We denote by $\Delta\latentVariableZ$ and $\Delta\jointTau$ the difference in the context vector and in the DeLaN torque with respect to the undisturbed run.

We evaluate 16 logarithmically spaced frequencies between 0.2~Hz and 24~Hz, remaining below the 25~Hz Nyquist frequency. At each frequency, we discard the first 2~s of each run and divide the remaining signal into periods of duration $1/f$. Consecutive periods are then aggregated into overlapping windows containing at least three periods and spanning at least 10~s. The windows are shifted by half their length, rounded down to a whole number of periods.

For each window, disturbed joint $j$, and context component $i$, we fit the fundamental harmonic by least squares:
\begin{equation}
\label{eq:complex_matrix_gain_element}
\Delta\latentVariableZ_{ij}(t) = \alpha_{ij} \sin (2\pi f t) + \beta_{ij} \cos( 2\pi f t) + c_{ij}.
\end{equation}
The resulting complex gain matrix of the context vector,
$\FrequencyGainMat_{\latentVariableZ} \in \mathbb{C}^{10\times7}$,
is obtained from the fitted harmonic coefficients as
\begin{equation}
    \label{eq:complex_matrix_gain}
    \FrequencyGainMat_{\latentVariableZ}(f)_{ij} = \frac{\alpha_{ij} + \mathrm{i}\beta_{ij}}{\SinusoidalAmp}.
\end{equation}
Since the encoder input is normalized by its training standard deviation, this matrix represents the frequency-dependent gain per standard deviation of the input disturbance. For each window, we compute the largest singular value $\sigma_\mathrm{max}(\FrequencyGainMat_{\latentVariableZ})$, and take its maximum over the windows of a run. The same procedure is performed for the residual joint torques $\jointTau$.

\section{Simulation Trajectory Tracking Results}
\label{app:tracking_results}
Tables~\ref{tab:sim_id_absolute} and~\ref{tab:sim_ood_absolute} give the median, mean and interquartile range of the metrics shown in the box plots of Fig.~(\ref{fig:all_boxplot_id_results}-\ref{fig:all_boxplot_od_results}) in Sec.~\ref{subsec:trajectory_trackin_sim}, over the 600 simulated runs, for the in-distribution and out-of-distribution payloads, respectively. The three controllers are the nominal MPC (Nominal),
CaDeLaC with the LSTM encoder and the input pre-filter (LSTM), and CaDeLaC with BoDeLaN (REN). Each run is summarized by the RMS over time of each metric, averaged over the seven joints. All runs are included, also the 3 runs per learned model with QP failures in the out-of-distribution evaluation.

\begin{table}[h]
\centering
\caption{Median, mean and interquartile range of the tracking metrics over 600 simulated runs (30 environments with in-distribution payloads, 20 trajectories each). The metrics of each run are averaged over the seven joints.}
\label{tab:sim_id_absolute}
\small
\begin{tabular}{lcccccc}
\hline
Model & $\bar{e}_\mathbf{q}$ & $\bar{e}_{\dot{\mathbf{q}}}$ & $\bar{e}_{\boldsymbol{\tau}}$ & RMS$(\ddot{\mathbf{q}})$ & RMS$(\dot{\boldsymbol{\tau}})$ & RMS$(\dot{\mathbf{z}}_\mathrm{n})$ \\
 & [rad] & [rad/s] & [Nm] & [rad/s$^2$] & [Nm/s] & [1/s] \\
\hline
\multicolumn{7}{c}{Median} \\
Nominal & 0.066 & 0.091 & 2.92 & \textbf{1.33} & \textbf{10.8} & -- \\
LSTM & \textbf{0.018} & 0.101 & 1.86 & 3.55 & 69.2 & \textbf{4.67} \\
REN & 0.029 & \textbf{0.090} & \textbf{1.55} & 2.41 & 47.3 & 4.77 \\
\hline
\multicolumn{7}{c}{Mean} \\
Nominal & 0.079 & \textbf{0.109} & 3.25 & \textbf{1.36} & \textbf{11.5} & -- \\
LSTM & \textbf{0.030} & 0.174 & 2.60 & 4.52 & 94.7 & 5.34 \\
REN & 0.040 & 0.133 & \textbf{2.42} & 3.16 & 62.9 & \textbf{4.82} \\
\hline
\multicolumn{7}{c}{Interquartile range} \\
Nominal & 0.066 & 0.075 & 3.26 & \textbf{0.22} & \textbf{4.6} & -- \\
LSTM & 0.015 & 0.158 & 1.37 & 1.94 & 47.1 & 2.03 \\
REN & \textbf{0.008} & \textbf{0.018} & \textbf{0.26} & 0.30 & 8.0 & \textbf{0.86} \\
\hline
\end{tabular}
\end{table}

\begin{table}[h]
\centering
\caption{Median, mean and interquartile range of the tracking metrics over 600 simulated runs (30 environments with out-of-distribution payloads, 20 trajectories each). The metrics of each run are averaged over the seven joints.}
\label{tab:sim_ood_absolute}
\small
\begin{tabular}{lcccccc}
\hline
Model & $\bar{e}_\mathbf{q}$ & $\bar{e}_{\dot{\mathbf{q}}}$ & $\bar{e}_{\boldsymbol{\tau}}$ & RMS$(\ddot{\mathbf{q}})$ & RMS$(\dot{\boldsymbol{\tau}})$ & RMS$(\dot{\mathbf{z}}_\mathrm{n})$ \\
 & [rad] & [rad/s] & [Nm] & [rad/s$^2$] & [Nm/s] & [1/s] \\
\hline
\multicolumn{7}{c}{Median} \\
Nominal & 0.200 & 0.261 & 10.06 & \textbf{1.67} & \textbf{20.7} & -- \\
LSTM & \textbf{0.065} & \textbf{0.237} & 4.83 & 3.98 & 94.0 & 5.08 \\
REN & 0.066 & 0.255 & \textbf{3.96} & 3.08 & 67.8 & \textbf{4.23} \\
\hline
\multicolumn{7}{c}{Mean} \\
Nominal & 0.207 & \textbf{0.265} & 10.10 & \textbf{1.74} & \textbf{21.3} & -- \\
LSTM & \textbf{0.072} & 0.272 & 13.35 & 6.48 & 121.1 & 5.35 \\
REN & 0.073 & 0.268 & \textbf{8.17} & 4.99 & 86.3 & \textbf{4.29} \\
\hline
\multicolumn{7}{c}{Interquartile range} \\
Nominal & 0.068 & 0.104 & 3.93 & \textbf{0.39} & \textbf{7.1} & -- \\
LSTM & 0.044 & \textbf{0.076} & \textbf{2.68} & 1.22 & 17.7 & 1.31 \\
REN & \textbf{0.032} & 0.167 & 2.87 & 1.13 & 41.2 & \textbf{0.73} \\
\hline
\end{tabular}
\end{table}
\FloatBarrier

\section{Hardware Pick-and-Place Task}
\label{sec:appendix_pp_task_desc}
To test the sensitivity of the encoders, a deterministic sinusoidal disturbance is added to the residual torque channel of the encoder input, before the low-pass filter, for joints $j = 1,\dots,5$:
\begin{equation*}
    d_j(t) = s_j \sum_{i=1}^{2} a_i \sin(2\pi f_i t + \phi_i),
\end{equation*}
with $(a_i, f_i, \phi_i) = (0.3,\,5\,\mathrm{Hz},\,0)$ and $(0.3,\,10\,\mathrm{Hz},\,0.3\,\mathrm{rad})$, and the scales $s = [1.6,\ 10,\ 10,\ 6.6,\ 1.0]$~Nm for joints 1 to 5. Joints 6 and 7 are not disturbed. The signal starts with the run and is sampled at the 50~Hz control rate. It changes only the input of the encoder; the measured joint states and torques are not modified. The same signal is used for the LSTM and the REN. Table~\ref{tab:real_robot_absolute} reports the results for the nomimal MPC, and CaDeLaC with both encoders with and without disturbance.

\begin{table}[h]
\centering
\caption{Real-robot pick-and-place results averaged over the seven joints.}
\label{tab:real_robot_absolute}
\small
\begin{tabular}{lcccccc}
\hline
Model & $\bar{e}_\mathbf{q}$ & $\bar{e}_{\dot{\mathbf{q}}}$ & $\bar{e}_{\boldsymbol{\tau}}$ & RMS$(\ddot{\mathbf{q}})$ & RMS$(\dot{\boldsymbol{\tau}})$ & RMS$(\dot{\mathbf{z}}_\mathrm{n})$ \\
 & [rad] & [rad/s] & [Nm] & [rad/s$^2$] & [Nm/s] & [1/s] \\
\hline
Nominal & 0.033 & 0.048 & 2.97 & \textbf{0.82} & \textbf{12.0} & -- \\
LSTM & \textbf{0.015} & 0.071 & 1.85 & 2.35 & 43.4 & 2.83 \\
REN & 0.026 & \textbf{0.042} & \textbf{0.97} & 1.13 & 17.4 & \textbf{1.44} \\
\hline
LSTM with Disturbance & \textbf{0.018} & 0.317 & 4.00 & 13.78 & 113.3 & 8.31 \\
REN with Disturbance & 0.025 & \textbf{0.055} & \textbf{1.31} & \textbf{1.70} & \textbf{33.6} & \textbf{2.26} \\
\hline
\end{tabular}
\end{table}

\end{document}